\documentclass{article}
\usepackage{spconf,amsmath,graphicx,hyperref,svg,gensymb}

\newcommand\blfootnote[1]{%
  \begingroup
  \renewcommand\thefootnote{}\footnote{#1}%
  \addtocounter{footnote}{-1}%
  \endgroup
}

\title{Thinking the Fusion strategy of multi-reference face reenactment}
\name{Takuya Yashima, Takuya Narihira, Tamaki Kojima}
\address{Sony Group Corporation}
\begin{document}
%
\maketitle
\begin{abstract}
In recent advances of deep generative models, face reenactment ---manipulating and controlling human face, including their head movement---has drawn much attention for its wide range of applicability. Despite its strong expressiveness, it is inevitable that the models fail to reconstruct or accurately generate unseen side of the face of a given single reference image. Most of existing methods alleviate this problem by learning appearances of human faces from large amount of data and generate realistic texture at inference time. Rather than completely relying on what generative models learn, we show that simple extension by using multiple reference images significantly improves generation quality. We show this by 1) conducting the reconstruction task on publicly available dataset, 2) conducting facial motion transfer on our original dataset which consists of multi-person's head movement video sequences, and 3) using a newly proposed evaluation metric to validate that our method achieves better quantitative results.
\end{abstract}

\vspace{-3mm}
\section{Introduction}
\label{sec:intro}
\blfootnote{©2022 IEEE. Personal use of this material is permitted. Permission from
IEEE must be obtained for all other uses, in any current or future media,
including reprinting/republishing this material for advertising or promotional
purposes, creating new collective works, for resale or redistribution to servers
or lists, or reuse of any copyrighted component of this work in other works.}Ever since neural network based generative models first showed its potential applicability for face generation, countless number of successive works have been published to this day. Generative models are able not only to generate non-existent faces, but also to edit face images. For example, poses, colors, parts, and even expressions of the faces in the image can be changed very naturally. 
Recently, a more advanced task where one person is \textit{manipulated} by another person's movement has emerged. In other words, the person is forced to make the same movement as another person does. This is called face reenactment or facial motion transfer and currently many researchers are interested in this topic for its wide applicability \cite{tolosana2020deepfakes}.
We refer the person to be manipulated as \textit{reference} person and the person whose motion is transferred as \textit{driving} person although they could be the same person, for example in reconstruction task.
Although purely 2D-based face reenactment models achieve impressive results, they still suffer from generation of unseen side of the face. We show that using multiple reference images with a proper feature fusion technique significantly improves the generation quality.
Contributions of this paper can be summarized as follows: 1) We propose a face reenactment model which can use multiple reference images to generate more faithful results.
2) We propose feature fusion methods which integrate incoming features properly to form a useful feature for a better generation result. 
3) We propose a novel evaluation metric specially designed for face reenactment. 

\vspace{-3mm}
\section{Related Work}
\label{sec:format}

\textbf{Face generation:} Human face has been a subject of interest in generative models even in its early work \cite{goodfellow2014generative}.
Recent models \cite{brock2018large, karras2019style} can generate high resolution images which are quite difficult to distinguish from real images even for human eyes.

\noindent
\textbf{Face edit:} StarGAN \cite{choi2018stargan} changes facial attributes like hair color, age, and even its gender. MaskGAN \cite{lee2020maskgan} can edit finer parts of the face with the aid of semantic segmentation. HoloGAN \cite{nguyen2019hologan} can change the pose of the given face images. Furthermore, expression is no exception. GANimation \cite{pumarola2018ganimation} can manipulate facial images with more controllability, and the generated images smoothly can change their expressions such as smiling and blinking.

\noindent
\textbf{Face reenactment:} While some works \cite{kim2018deep, head2head2020} utilize 3D morphable models for this task, 2D-based models are still actively explored. 
ReenactGAN \cite{Wu_2018_ECCV} is an early face reenactment model based on 2D image-to-image translation.
X2Face \cite{wiles2018x2face} proposed a model which internally uses geometric warp to transform input images. Similar ideas can be found in recent reenactment models such as First Order Motion Model \cite{Siarohin_2019_NeurIPS}, which uses additional decoder refining incoming image features and tricks to mitigate identity gaps when reference and driving persons differ.

\noindent
\textbf{Evaluation Metrics:} FID \cite{heusel2017gans} is one of the popular evaluation metrics for generative models including face reenactment \cite{brock2018large, karras2019style, zhang2020freenet, wang2021one}. However, simply using FID is not suited for the evaluation of face reenactment. FID just measures a distance of data distributions between true images and generated images and doesn't consider poses and angles of the faces.

\begin{figure*}[th]
  \centering
  \includegraphics[width=0.85\linewidth]{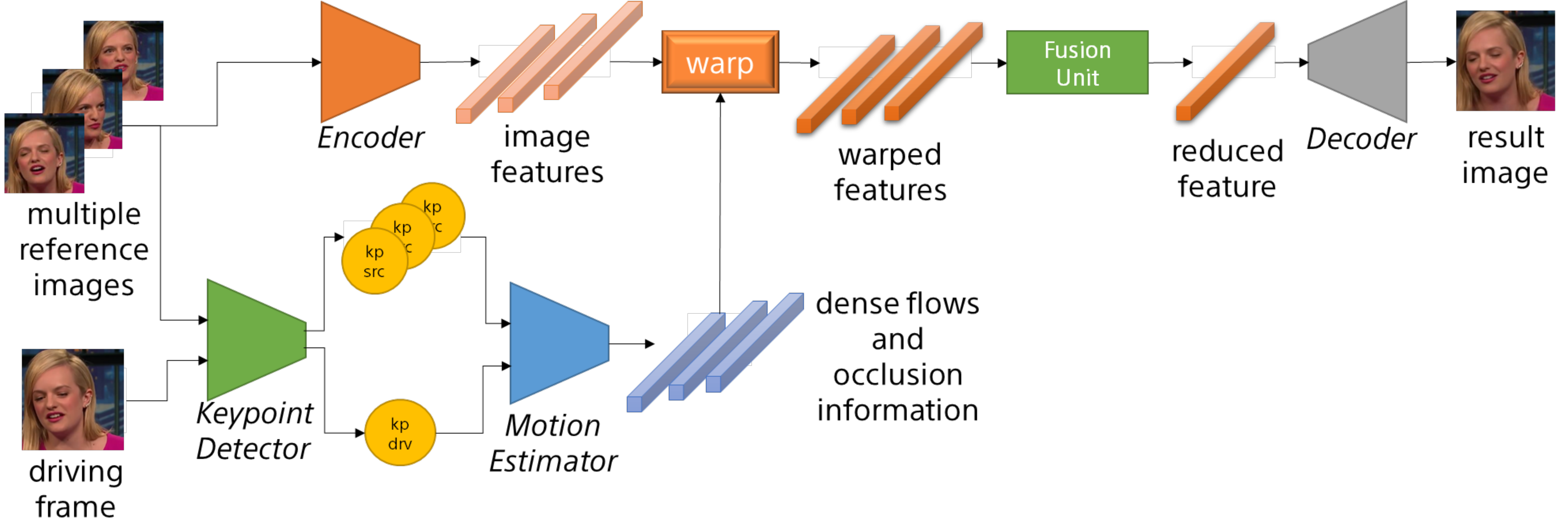}
  \caption{Architecture of our model when $K = 3$.}
  \label{fig:mfomm_arch}
  \vspace{-2mm}
\end{figure*}


\vspace{-3mm}
\section{Model}
\label{sec:pagestyle}


\vspace{-3mm}
\subsection{Model Overview}\vspace{-2mm}
\label{ssec:overview_fomm}

Our model is based on First Order Motion Model (abbreviated as FOMM) \cite{Siarohin_2019_NeurIPS}. 
It consists of 4 major networks; encoder, keypoint detector, motion estimator, and decoder. First, the encoder extracts features from multiple reference images and the keypoint detector predicts keypoints of the reference and the driving images. Based on these keypoints, the motion estimator predicts motion flow and occlusion map used for warping. These intermediate representations are then used to warp the image feature obtained by the encoder. This warping works as pose alignment, forcing the reference face to "pose" the same as the one in the driving image. The original FOMM takes the single warped feature to the decoder to output the result image whose head poses and expressions should look like driving images. We extend this model architecture by introducing the fusion unit to accept $K$ reference images as shown in Figure~\ref{fig:mfomm_arch}.
The rest of the architecture is the same as that of the original FOMM, including loss functions and constraints.
The fusion unit will extract appropriate features from each of the incoming features and combine them to form a fused feature. 
Details are described in the subsequent subsection.

\vspace{-3mm}
\subsection{Fusion methods}\vspace{-2mm}
\label{ssec:fusion_unit}
\textbf{Patch-wise weighted sum:} Our fusion method uses attention mechanism also utilized in GANimation \cite{pumarola2018ganimation}.
Inside the fusion unit, convolution with shared weights is applied to each of $K$ warped features and outputs single channel features.
Then, all features are stacked along a new dimension and softmax is applied to form the \textit{patch-wise masks}. In this mask map, the sum of values at the same pixel location will be 1.
These mask maps and warped features are then multiplied and summed up to form a fused feature. Table~\ref{patch_wise_mask} shows the effect of this method and it is clear that these masks filter out irrelevant features.

\begin{figure}[t]
  \centering
  \includegraphics[width=0.90\linewidth]{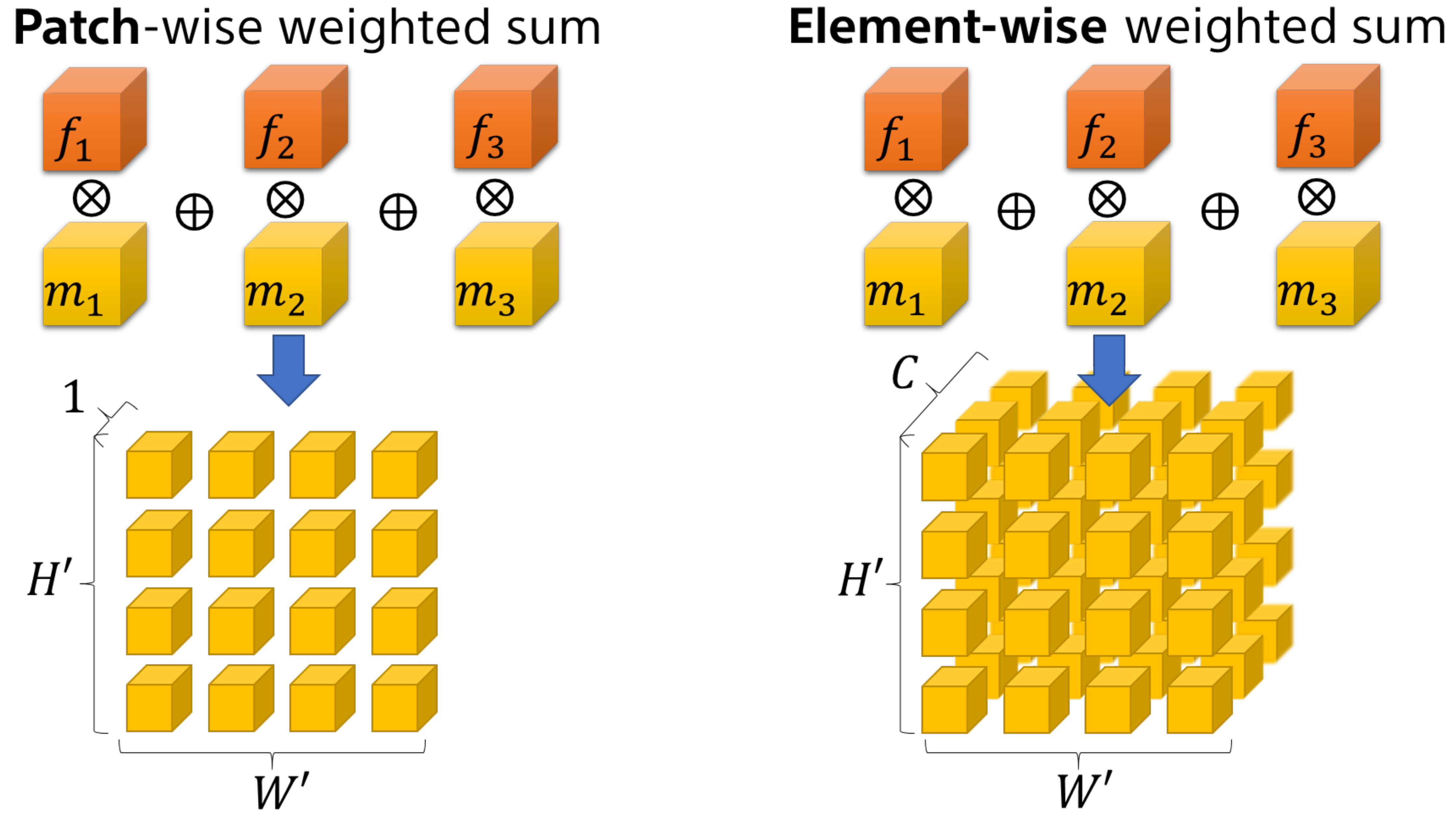}
  \caption{Difference between masks used for patch-wise and element-wise weighted sum fusion, when the number of references $K = 3$. Top: $f_{1}$, $f_{2}$ and $f_{3}$ are warped features shape of $(C, H', W')$. $m_{1}$, $m_{2}$, and $m_{3}$ are \textit{mask} maps. Bottom: A detailed view of the mask map. Each yellow block represents an individual weight of the masks.}
  \label{fig:patch_and_elem}
  \vspace{-2mm}
\end{figure}

\noindent
\textbf{Element-wise weighted sum:} This method is similar to the patch-wise weighted sum described above. The difference lies in the granularity of the mask. The mask map consists of multiple channels, where the number of channels matches that of the warped features. Thus, for each channel, different weights are stored in every pixel location. Figure~\ref{fig:patch_and_elem} shows how patch-wise and element-wise weighted sum fusions differ. We can expect this element-wise weighted sum fusion to convey the most fine-grained information from each of the warped features.


\begin{table}[h]
  \begin{tabular}{|c|c|c|c|}
    \hline \rm driving & \rm ref0 & \rm ref1 & \rm ref2 \\ \hline
    \hline
    \begin{minipage}{17mm}
      \centering
      \scalebox{0.18}{\includegraphics{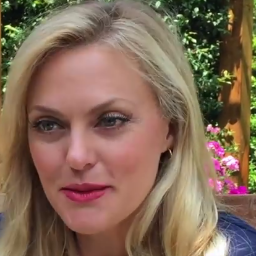}}
    \end{minipage} &
    \begin{minipage}{17mm}
      \centering
      \scalebox{0.18}{\includegraphics{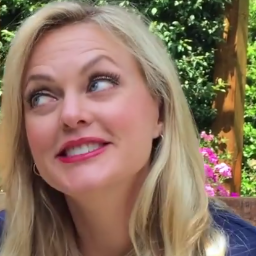}}
    \end{minipage} &
    \begin{minipage}{17mm}
      \centering
      \scalebox{0.18}{\includegraphics{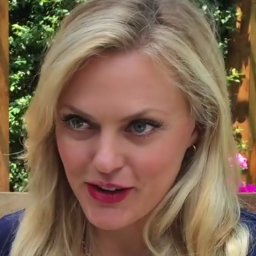}}
    \end{minipage} &
    \begin{minipage}{17mm}
      \centering
      \scalebox{0.18}{\includegraphics{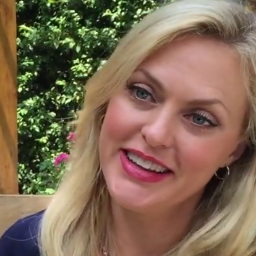}}
    \end{minipage} \\ \hline

    \hline
    \begin{minipage}{17mm}
      \centering Decoded warped feature
    \end{minipage} &
    \begin{minipage}{17mm}
      \centering
      \scalebox{0.18}{\includegraphics{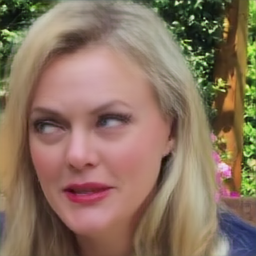}}
    \end{minipage} &
    \begin{minipage}{17mm}
      \centering
      \scalebox{0.18}{\includegraphics{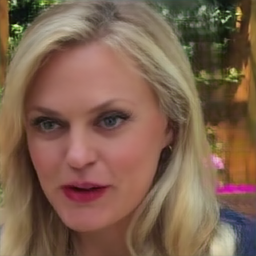}}
    \end{minipage} &
    \begin{minipage}{17mm}
      \centering
      \scalebox{0.18}{\includegraphics{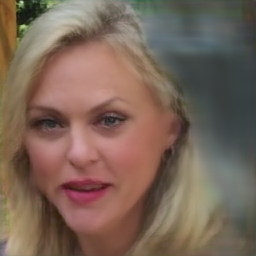}}
    \end{minipage} \\ \hline

    \hline
    \begin{minipage}{17mm}
      \centering Masked results 
    \end{minipage} &
    \begin{minipage}{17mm}
      \centering
      \scalebox{0.18}{\includegraphics{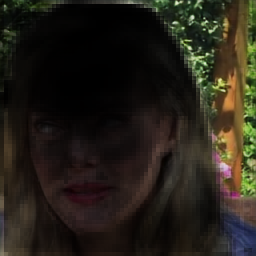}}
    \end{minipage} &
    \begin{minipage}{17mm}
      \centering
      \scalebox{0.18}{\includegraphics{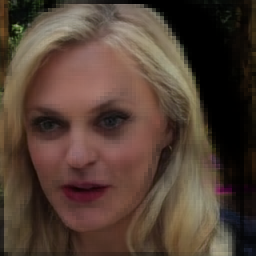}}
    \end{minipage} &
    \begin{minipage}{17mm}
      \centering
      \scalebox{0.18}{\includegraphics{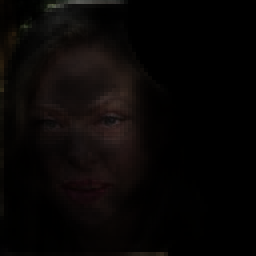}}
    \end{minipage} \\ \hline

  \end{tabular}
  \caption{Effect of patch-wise weighted sum based fusion. From top to bottom: input images, decoder's outputs given each warped features made from the image in the same column, and images after applying the mask.}
  \label{patch_wise_mask}
\end{table}

\vspace{-3mm}
\section{Experiments}
\label{sec:typestyle}

\vspace{-3mm}
\subsection{Task definitions}\vspace{-2mm}
\label{ssec:task_description}

\noindent
\textbf{Reconstruction:} We focus on how the model generates a video sequence given the first frame of the video as reference and other frames as driving frames. It can be considered as self-reenactment since driving and reference person are the same. Ideally, after combining the resulting images to form a video sequence, the generated video sequence should exactly look identical to the driving video sequence. This is frequently adopted as a basic evaluation metric for face reenactment model \cite{wiles2018x2face, Siarohin_2019_NeurIPS}.

\noindent
\textbf{Motion transfer:} Given a reference image of an individual and driving frames of other individual or environments, motion transfer is a task which manipulates the face / head of the reference by the movements and the facial expression of the driving frames.
The generated sequences are expected to follow the movements and the facial expression of the driving frames.

\vspace{-3mm}
\subsection{Dataset}\vspace{-2mm}
\label{ssec:dataset}
\textbf{VoxCeleb1:} VoxCeleb1 \cite{nagrani2017voxceleb} is a dataset containing a large amount of video sequences of human speech, originally uploaded to Youtube. We followed FOMM \cite{Siarohin_2019_NeurIPS} for obtaining and pre-processing the data. We used this data for training the model and for evaluating the reconstruction task.

\noindent
\textbf{Internal Data:} For evaluation of motion transfer, we used our internal data. Though the data size is very small compared to VoxCeleb1, it contains videos of the predetermined head movements of 5 different subjects, under 2 types of illumination (except 1 subject with only 1 illumination type, making total of 9 patterns). All the subjects are equipped with a device which records head angle (yaw, pitch, roll) and instructed to move their head in several predetermined ways. The data are then split into sequences of; 1) yaw-varying, 2) pitch-varying and 3) roll-varying for evaluation of the motion transfer task.
Note that the speed of head movements varies, so these sequences are not temporally aligned.

\vspace{-3mm}
\subsection{Implementation}\vspace{-2mm}
\label{ssec:imple}
We have implemented the model with Neural Network Libraries \cite{narihira2021neural}.
Following FOMM \cite{Siarohin_2019_NeurIPS}, we have trained our model using VoxCeleb1 \cite{nagrani2017voxceleb}. 
The difference from FOMM in the training  is that $K$ frames are sampled instead of 1 for the reference frames. Note that reference frames and the driving frames were randomly sampled. We used $K = 3$ in all of our experiments.
Training consists of 2 stages, the first for 100 epochs without discriminator and the second for 50 epochs with discriminator. We used batch size of 32, initial learning rate of 0.0002 with linear decay by 0.1 at epoch 60 and 90, and used Adam \cite{kingma2014adam} as our optimizer. We use 4 NVIDIA A100 GPUs for training.

\vspace{-3mm}
\section{Evaluation}
\label{sec:eval}


\vspace{-3mm}
\subsection{Reconstruction}\vspace{-2mm}
\label{ssec:eval_reconstruction}

Following ~\ref{ssec:task_description}, once we generate the video sequences, we calculate L1 distance (L1D) between real and generated sequences. Reported scores are averaged L1D over all the samples. We use test set of VoxCeleb1 \cite{nagrani2017voxceleb} as driving videos, and perform this reconstruction task in the same manner, except that we use 3 reference images as multiple references. Unlike in training, these 3 images are always 1) the first frame, 2) the last frame, and 3) the middle frame of the driving video. This ensures that we do not impose any preference on appearance such as pose and expression. This is not an optimal usage of our model, but we employed for simplicity and fairness. For fair comparison, we also calculate Average Keypoint Distance (AKD) and Average Euclidean Distance (AED) as in FOMM \cite{Siarohin_2019_NeurIPS}. 
We compare these results with 2 baseline models 1) original FOMM, and 2) pseudo multi-reference FOMM. The latter uses the same model as the former, but we run the same model 3 times with 3 reference images and obtain 3 result images. We then pick the \textit{best} result image closest to the corresponding driving image. We used L1D for the distance comparison. This gives an advantage to the model in terms of accuracy, and it should lead to better result than original FOMM. Selection strategy for reference images for this pseudo multi-reference model is the same as the one used for our model (the first, the last, and the middle frame of the video of interest).

\noindent
\textbf{Result:} Table~\ref{table:recon_table} shows a summary of the result. Our model performs better than the pseudo multi-reference model, which virtually uses the ground truth information when choosing the frames. 

\begin{table}[tbhp]
  \centering
  \begin{tabular}{l||c|c|c}
    \hline
    Metrics & L1D  &  AKD  &  AED  \\
    \hline \hline
    FOMM & 0.043 & 1.294 & 0.140 \\
    (Pseudo multi-ref) FOMM & 0.033 & 1.191 & 0.093 \\
    Ours(Patch-wise) & \textbf{0.031} & \underline{1.159} & \underline{0.084} \\
    Ours(Element-wise) & \underline{0.032} & \textbf{1.156} & \textbf{0.078} \\
    \hline
  \end{tabular}
  \caption{Quantitative Evaluation of Reconstruction  Task on VoxCeleb1.}
  \label{table:recon_table}
\end{table}

\vspace{-3mm}
\subsection{Motion transfer}\vspace{-2mm}
\label{ssec:eval_motion_transfer}

Motion transfer is difficult to evaluate, especially in a quantitative manner. Unlike the reconstruction task where it is easy to obtain ground truth images, it is very difficult to obtain the ground truth images for motion transfer unless we have predefined movements or patterns. Therefore, we have evaluated this task on our internal data. We have generated the videos for every possible pair among 9 different patterns(except the same person under the same illumination). Reported scores are averaged over these pairs. As for the references, we used 3 images which are frontal, left-side (yaw $-30{\degree}$), and right-side (yaw $+30{\degree}$) faces, no matter which sequence to be evaluated.

\noindent
FID \cite{heusel2017gans} is one of the metrics frequently used for the evaluation of generative models including motion transfer task \cite{zhang2020freenet, wang2021one}.
However, it cannot adequately measure the quality of face reenactment. Motion transfer shall be evaluated in terms of how accurately the generated movement follows the driving head movement.
Therefore, we propose a novel evaluation metric utilizing binned angles as described below. 

\begin{figure}[t]
  \centering
  \includegraphics[width=0.85\linewidth]{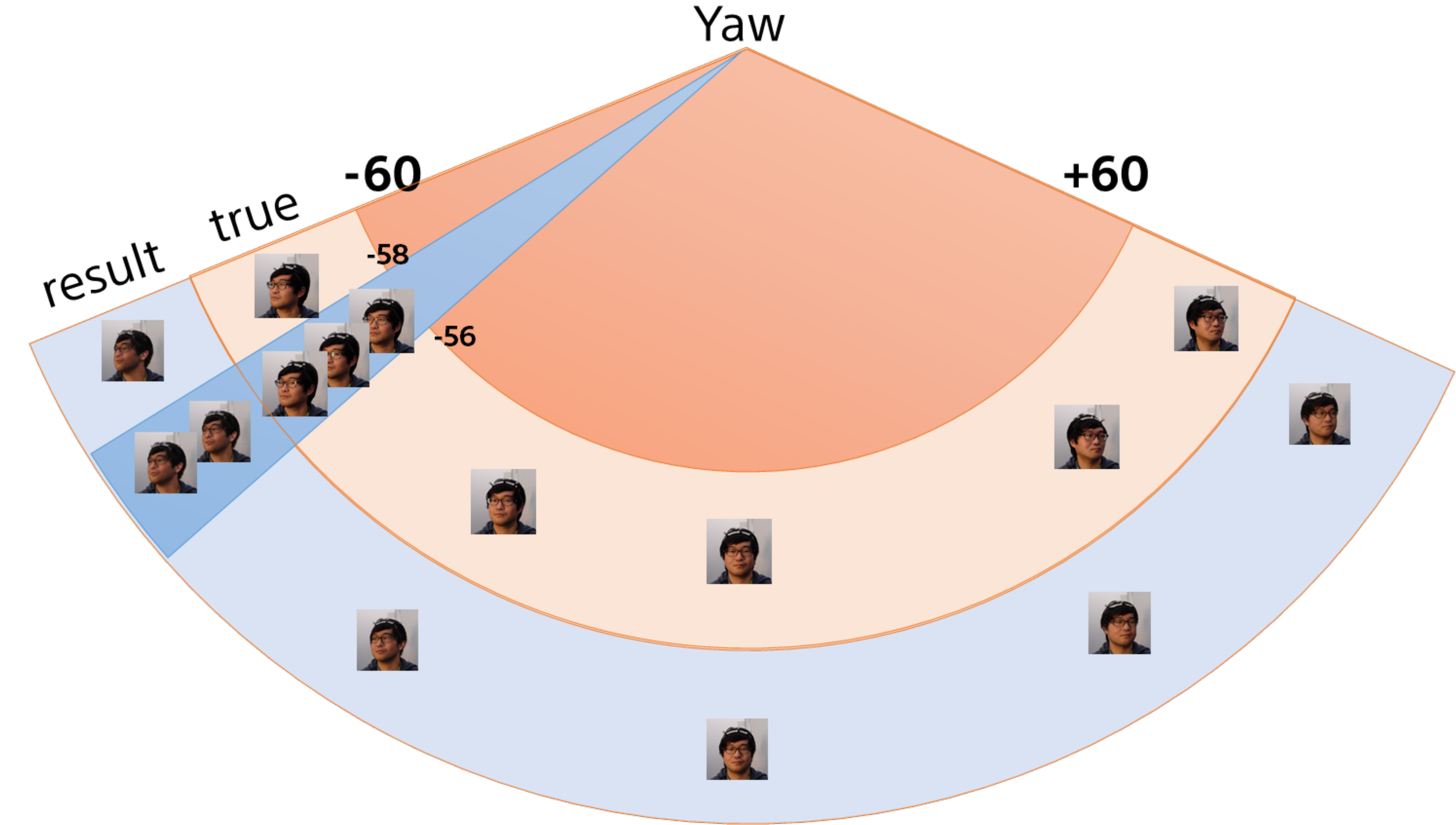}
  \caption{Bin-based evaluation.}
  \label{fig:bin_eval}
  \vspace{-4mm}
\end{figure}

\noindent
\textbf{Bin-based angle-wise evaluation:} As described in \ref{ssec:dataset}, our original data for evaluation come with the head angle annotation per frame. Suppose we generate one result image with a driving image of person A with reference images of person B, then we would get a result image of person B whose pose and head angle should be the same as that of person A. 
In other words, pose and head angle in driving and result images should match per frame. Therefore, by referring to the angle information of driving person, we can annotate angle information to each of result images. 

\noindent
Next, we split the angular ranges into multiple bins. For example, we split angular range of yaw axis from $+60{\degree}$ to $-60{\degree}$ into multiple bins on every 2 degrees with no overlap, as shown in Figure~\ref{fig:bin_eval}.
Then we can compare images within each bin since they should look alike.
For comparison, we use LPIPS \cite{zhang2018perceptual} which measures the perceptual similarity between semantically similar images, and is more reliable \cite{bang2020discriminator} than FID. 
When comparing images within one bin, we have to consider the case where the numbers of true images and result images that fall to that bin do not match.
For example, as shown in Figure~\ref{fig:bin_eval},
we have 3 true images that fall to the bin $[-58{\degree}, -56{\degree}]$, while we have only 2 result images that fall to the same bin. In this case, we simply average the features of true images and result images, respectively. These averaged features are regarded as  representative features of that bin. If there is no true or result image, we simply ignore that bin for calculation. Once we get representative features for both true and result images, we can calculate LPIPS scores per bin, following its definition.
Reported scores are averaged over bins.
So, the representative feature at $l$th layer $y^{l}$ can be obtained by 

\vspace{-6mm}

\begin{equation}
y^{l} = \frac{1}{N}\sum_{n=1}^{N}\mathcal{F}(I_{n}), \\ \label{eqn:1}
\end{equation}

where $N$ is the number of images in a bin of interest, $\mathcal{F}$ is a feature extractor, AlexNet \cite{krizhevsky2012imagenet}, up to the $l$th layer, and $I$ is either true or result images. 
Also, taking representative features like this allows us to use other evaluation metrics which utilize image features, such as face embedding.

\noindent
\textbf{Result:} We calculated the score for each sequence. As shown in Table~\ref{table:motion_table}, our element-wise fusion model performs the best in all sequences. It is worth noting that it achieves the best score not only in yaw-varying sequence but also in other sequences, despite the reference images varying only in yaw angle. Note that we only report the result with original FOMM and not with the pseudo multi-reference FOMM used in Reconstruction task, since it requires the ground truth to choose the best result. 

\begin{table}
  \centering
  \begin{tabular}{l||c|c|c}
    \hline
    Metrics & \multicolumn{3}{c}{LPIPS} \\
    \hline
    sequence & yaw & pitch & roll \\
    \hline \hline
    FOMM   & 0.117 & 0.131 & 0.160 \\
    Ours(Patch-wise)   & \underline{0.086} & \underline{0.108} & \underline{0.142} \\
    Ours(Element-wise) & \textbf{0.085} & \textbf{0.107} & \textbf{0.141} \\
    \hline
  \end{tabular}
  \caption{Quantitative evaluation of motion transfer on internal data.}
  \label{table:motion_table}
  \vspace{-2mm}
\end{table}

\vspace{-3mm}
\section{Conclusion}
\label{sec:conclusion}
\vspace{-3mm}
In this paper, we proposed a multi-reference face reenactment model which is a simple extension yet turns out to be effective. Quantitative evaluation shows that our model performs better than simply using multiple face reenactment models, which implies the internal feature fusion unit actually works for integrating important information from incoming image features. Since the internal network architecture is independent of how many reference images are used, once the model is trained, users can give arbitrary number of reference images and we can expect that the more reference users give, the better the result would be. We also propose a new evaluation metric designed for motion transfer task, which has been difficult to evaluate quantitatively. 
Most of the previous works rely on FID or qualitative evaluation only \cite{kim2018deep, zhang2020freenet, wang2021one}.
Thus, we hope our bin-based angle-wise evaluation metric help evaluate the face reenactment models in a different light.





\vspace{-3mm}
\bibliographystyle{IEEEbib}
\bibliography{refs}

\end{document}